# Dew Point modelling using GEP based multi objective optimization

Siddharth Shroff[#1], Vipul Dabhi[*2]

[#] *I.T Department, Dharmsinh Desai University*
*Nadiad,*
*India*

[1]siddharthshroff61@yahoo.com

[*] *I.T Department, Dharmsinh Desai University*
*Nadiad,*
*India*

[2] vipul.k.dabhi@gmail.com

*Abstract*—**Different techniques are used to model the relationship between temperatures, dew point and relative humidity. Gene expression programming is capable of modelling complex realities with great accuracy, allowing at the same time, the extraction of knowledge from the evolved models compared to other learning algorithms. We aim to use Gene Expression Programming for modelling of dew point. Generally, accuracy of the model is the only objective used by selection mechanism of GEP. This will evolve large size models with low training error. To avoid this situation, use of multiple objectives, like accuracy and size of the model are preferred by Genetic Programming practitioners. Solution to a multi-objective problem is a set of solutions which satisfies the objectives given by decision maker. Multi objective based GEP will be used to evolve simple models. Various algorithms widely used for multi objective optimization, like NSGA II and SPEA 2, are tested on different test problems. The results obtained thereafter gives idea that SPEA 2 is better than NSGA II based on the features like execution time, number of solutions obtained and convergence rate. We selected SPEA 2 for dew point prediction. The multi-objective base GEP produces accurate and simpler (smaller) solutions compared to solutions produced by plain GEP for dew point predictions. Thus multi objective base GEP produces better solutions by considering the dual objectives of fitness and size of the solution. These simple models can be used to predict future values of dew point.**

## I. INTRODUCTION

Gene Expression Programming (GEP) is an evolutionary algorithm that finds the model which could satisfy the training data dependencies. Other machine learning algorithms like Artificial Neural Networks can produce black box models. Moreover the solutions found by ANN and SVM in search space are not diverse in ANN and Support Vector Machines (SVM) as compared to evolutionary algorithms. Due to lack of prior knowledge required to model a relation, the best possible way to deal with this issue is using evolutionary based algorithms. Multi objective optimization is an area of multiple criteria decision making that is concerned with mathematical optimization problems involving more than one objective function to be optimized simultaneously. Using GEP to model the forecasting model of Dew point along with the optimization of that model using multiple objectives is the aim of the research work.

## II. BACKGROUND

### A. *Role of multi objective optimization*

Multi objective optimization is an area that deals with decisions making process where more than one objective leads to the preferred accurate solution. Multi objective optimization is applied in many fields where optimal decisions are to be taken considering two or more conflicting objectives. Thus here we are concerned with optimizing mathematical models involving more than one objective function to be optimized simultaneously. An obvious conflicting objectives in optimizing the mathematical equation can be the complexity of the equation and error less output .Using GEP modelling tool dew point formulae are generated, but in order to validate the formula based on its accuracy, complexity of any other optimizing objective we need multi-objective optimization.

Multi-objective optimization problem have set of solutions as the final output of GEP run. These solutions are obtained based on the trade-off between conflicting objectives. There exists large number of solutions satisfying the trade-off between conflicting objectives, known as Pareto solutions. A solution is called non-dominated if and only if none of the objective functions can be improved in value without degrading some of the other objective values. This set of non-dominated solutions is known as Optimal Pareto front. The goal may be finding a set of Pareto optimal solutions, and quantifying the trade-offs in satisfying the different objectives.

### *B. Introduction to multi objective algorithms*

Survey of different research papers based on multi-objective optimization gave an idea of recently used algorithms and their pros and cons. These algorithms are divided into two groups based on use of elitism. Recently invented algorithms like NSGA II, SPEA, PAES, and SPEA II etc. use elitism. Out of these SPEA and SPEA II uses external population which stores the best results obtained so far. Elitism is used to preserve the previous best results and thus have a better search ability in solution space. Below we describe well known multi objective algorithms in brief:

#### *1) Improved version of Non dominated Sorting Genetic Algorithm(NSGA II)*

This algorithm sorts the solutions according to their non-dominance based on the objectives with the other solutions in the space. Earlier version of NSGA performs following steps: In order to find the first non-dominated pareto front for processing population of size N, for M number of objectives, each individual is compared to other individual which takes time O ($MN^2$). The same procedure is repeated for generating second pareto front in worst case and same procedure for N pareto fronts, leading to time complexity of O ($MN^3$). The space complexity for this algorithm is O (N) [12].

Non dominated Sorting Genetic Algorithm had many issues like – High time complexity O ($MN^3$), lack of elitism, absence of separate sharing parameter [15]. Thus to solve these issues in NSGA a new improved version is introduced which outperforms algorithms like SPEA, PAES using features like maintaining diversity and convergence to optimal Pareto front.

New version of NSGA introduces two parameters in addition to above algorithm - domination count $nd_p$ i.e. number of solutions which dominate solution p and $dS_p$, a set of solutions that the solution p dominates. Now, for each solution $p$ with $nd_p = 0$ i.e. the best solutions for which we have to visit each member ($q$) of its set $dSp$ and reduce its domination count by one. During this iteration, if for member $q$ the domination count becomes zero, it is copied in a separate list Q. Now, the above procedure is continued with each member of Q till all fronts are identified [15].The overall complexity of the algorithm is O ($MN^2$), governed by the non-dominated sorting part of the algorithm.

#### *2) Strength Pareto Evolutionary Algorithm(SPEA)*

SPEA introduces the idea of maintaining external population which was not there in NSGA or NSGA II. This external archive stores all non-dominated solutions obtained so far after each generation. Here a new fitness calculation method is used which assigns fitness based on the number of solutions they dominate in the external population as well as current population. This fitness is known as strength of that solution.

For each solution, a strength value is defined as ratio of number of solutions dominated by the solution whose strength is to be found to the total number of solutions in the population. Finally, the rank of the solution becomes the addition of the strengths of all the solutions which are dominated plus one.

Each individual is assigned strength value depending upon the solutions they dominate in the mixed population set of archived and current population. Therefore, a wide, uniformly distributed set of non-dominated solutions is encouraged.

Some problems have large pareto optimal solutions In order to reduce the number of solutions from the optimal Pareto front SPEA uses clustering technique to reduce the solution size.

#### *3) Improved version of SPEA(SPEA II)*

The main differences between SPEA2 and SPEA are [14]:

- A new fitness assignment scheme is used. This scheme takes into account, number of individuals each it dominates and it is dominated by.
- A nearest neighbor density estimation technique (k-NN) is incorporated which allowing more precise guidance of the search process compared to clustering technique used in SPEA
- An archive truncation method preserves boundary solutions.
- Archive size is fixed as compared to SPEA.

Strength of solutions includes comparisons with that of archive and population. SPEA2 considers fitness calculation for each individual both dominating and dominated solutions. Both dominating and dominated solutions are taken into account for comparison.

Algorithm may fail when most individuals do not dominate each other. Additional density information is added to differentiate between individuals having identical raw fitness values. Here the density of a solution is an inverse function of the distance to the k -th nearest data point of that solution. SPEA II inverses the distance to the k –th nearest neighbour as the density estimate. [14]

SPEA II considers the final fitness as the sum of raw fitness gained by adding up the strength values of every solution that it dominates in population set as well as archive set and the density factor. Thus final fitness assignment is sum of the raw fitness obtained and the density factor.

The run-time of the fitness assignment procedure is dominated by the density estimator (O ($M^2$ log M)), while the calculation of the S and R values is of complexity O ($M^2$), where M=N+N'. (N=current population, N'= archived population).

### *C. Comparison of multi objective algorithms*

The detailed comparison of different algorithms based on valuable features are given below

**TABLE 1:** COMPARISON OF MULTI OBJECTIVE ALGORITHMS

| **Algorithm** | **Fitness Assignment** | **Selection** | **Diversity Mechanism** | **Elitism** | **External Population** | **Time Complexity** | **Tool kit Support (JCLEC/ ECJ)** |
|---|---|---|---|---|---|---|---|
| Fast Non-dominated Sorting Genetic Algorithm (NSGA-II) | Non-dominated sorting ranking | Tournament Selection | Niching by Crowded distance | Yes | No | O($M^2$) | Yes |
| Strength Pareto Evolutionary Algorithm(SPEA) | Strength calculated from dominated solutions | Sharing of fitness function | Clustering Technique | No | Yes | O($M^2$) M=N+ N'.Population + Archive pop. | Yes |
| SPEA2 | Strength of dominating and dominated solutions | Tournament Selection | K-NN | Yes | Yes | O($M^2$log M) | Yes |

## III. EXPERIMENTAL METHODOLOGY AND RESULTS

There are many toolkits available for evolutionary computing. ECJ (Evolutionary Computing in Java), JCLEC, Lil-GP, EGIPSYS, GeneXpro, Watchmaker, Open Beagle etc. are examples of toolkits. Amongst all of the available toolkit ECJ, JCLEC, Lil-GP, EGIPSYS and Open Beagle are open source and freely available. ECJ and JCLEC are implemented in java language and thus are platform independent.

ECJ has more features compared to other toolkit and reduces the effort of implementing any feature from scratch. Most importantly GEP is supported in ECJ. In order to test the toolkit and its performance many test problems are taken based on complexity (size) of the problem and usage of constants etc.

### *A. Basic steps of execution of GEP in ECJ*

The following steps are performed for execution of GEP in ECJ

- ECJ version 18 supports GEP
- Compile the GEP files available in ECJ/ec/gep directory.
- Generate java file as per requirement in application. Compile the java file. Also generate parameter file specifying the parameters used by GEP.(Table 2, Table 3 & Table 4)
- Execute the application by using
  java ec.Evolve –file ..//.\ecj\ec\app\gep\Dew\dew.params
- The parameter file also specifies the statistics file to be generated at the end of execution. In order to generate the statistics SimpleShortStatistics.java and GEPSimpleStatistics.java files are used.

#### *1) Test Problem I :*

We have used following expression as test problem 1

y= cos(sqrt(sin(c))*cos(b)*sin(a) + tan(d-e) …………..[1]

The problem is considered by [8] and solved using GEP. The parameters to be used in GEP and changes in the parameter file are explained below. The run for 1000 generations with sub population size of 100, gene head size of 8 and number of genes equals to 3. The fitness function used is Root Relative Squared Error.

**TABLE 2**: GENERAL PARAMETERS

| General Parameters | Changes in parameter file |
|---|---|
| Number of genes | gep.species.numgenes = 3 |
| Gene Head Size | gep.species.gene-headsize = 8 |
| Size of sub population | pop.subpop.0.size =100 |

With function set of size 9, the following functions are added and shown in Table 3 with their weights and symbols used.

**TABLE 3**: FUNCTIONS SET USED

| Function Name | Weight | Symbol | Changes in parameter file |
|---|---|---|---|
| Addition | 1 | + | gep.species.symbolset.function.0 = Add |
| Multiplication | 1 | * | gep.species.symbolset.function.0 = Mul |
| Subtraction | 1 | - | gep.species.symbolset.function.0 = Sub |
| Division | 1 | / | gep.species.symbolset.function.0 = Div |
| Exponential | 1 | Exp | gep.species.symbolset.function.0 = Exp |
| Sin | 1 | Sin | gep.species.symbolset.function.0 = Sin |
| Cos | 1 | Cos | gep.species.symbolset.function.0 = Cos |
| Tan | 1 | Tan | gep.species.symbolset.function.0 = Tan |
| Sqrt | 1 | Sqrt | gep.species.symbolset.function.0 = Sqrt |

GEP has variety of genetic operators compared to GP. It also has operators for deciding whether to use the constants or not in the final solution. Domain range of the constants has to be specified by the user. Mutation, crossover inverse transposition operations occurs with the specified probability ($D_c$).Table 4 specifies the different probabilities of these genetic operators.

**TABLE 4:** GENETIC OPERATORS PROBABILITY

| Genetic Operators | Probability | Changes in parameter file |
|---|---|---|
| Inversion | 0.1 | gep.species.inversion-prob=0.1 |
| Mutation | 0.044 | gep.species.mutation-prob=0.3 |
| IS transposition | 0.1 | gep.species.istransposition-prob=0.1 |
| RIS transposition | 0.1 | gep.species.ristransposition-prob=0.1 |
| 1-pointrecombinatation | 0.3 | gep.species.onepointrecomb-prob=0.3 |
| 2-pointrecombinatation | 0.3 | gep.species.twopointrecomb-prob=0.3 |
| Gene recombination | 0.1 | gep.species.generecomb-prob=0.1 |
| Gene transposition | 0.1 | gep.species.genetransposition-prob=0.1 |
| RNC-mutation | 0.01 | gep.species.rnc-mutation-prob=0.01 |
| $D_c$ –mutation | 0.044 | gep.species.dc-mutation-prob=0.044 |
| $D_c$ –inversion | 0.1 | gep.species.dc-inversion-prob=0.1 |
| $D_c$ -IS transposition | 0.1 | gep.species.dc-istransposition-prob=0.1 |

The best model obtained by GEP is given in equation [2]

(((C-D)+cos((sqrt(cos(D))*sin(D))))+sqrt((B-E))) ……………[2]

Here the size of the evolved solution is 14 and the fitness is 0.0 .

*B. Solving problem using multi objective GEP*

Fitness and size of expression tree are two objectives considered while solving problem in multi objective GEP environment. Low fitness and least size of the equation is considered as the best solution.

Along with the number of objectives it has to be specified that the objectives are to be minimized or maximized i.e. minimizing the value gives better solution or maximizing the value gives better solutions. Here both the objectives are to be minimized to attain the desired solution. Thus the following changes are to be made in parameter file to minimize the objective values.

**pop.subpop.0.species.fitness.maximize = false**

Following additional parameters need to be specified for multi objective environment as shown in Table 4.

**TABLE 5:** MULTI OBJECTIVE PARAMETERS

| Parameter | Value | Changes in parameter file |
|---|---|---|
| Number of objectives | 2 | multi.fitness.num-objectives = 2 |
| Minimum value of 1st objective | 0.0 | multi.fitness.min.0 = 0 |
| Maximum value of 1st objective | 1.0 | multi.fitness.max.0 = 1 |
| Minimum value of 2nd objective | 4 | multi.fitness.min.1 = 4 |
| Maximum value of 2nd objective | 64 | multi.fitness.max.1 = 64 |

Using the fitness function of NSGA II and tournament size of 2, the optimal pareto front obtained is shown in Figure 1. Executing every test problems for 30 runs and collecting all these individuals are plotted on scatter plot and the optimal pareto front is shown by solid line. The parameters to be specified for SPEA 2 are shown in the Table 6

**TABLE 6**: PARAMETERS FOR NSGA II

| Parameter in NSGA II | Changes in parameter file |
|---|---|
| Selection of parent parameter file | parent.0=../../../multiobjective/nsga2/nsga2.params |
| Usage of fitness function of NSGA II | pop.subpop.0.species.fitness = ec.multiobjective.nsga2.NSGA2MultiObjectiveFitness |
| Selection of Tournament Size | select.tournament.size= 2 |

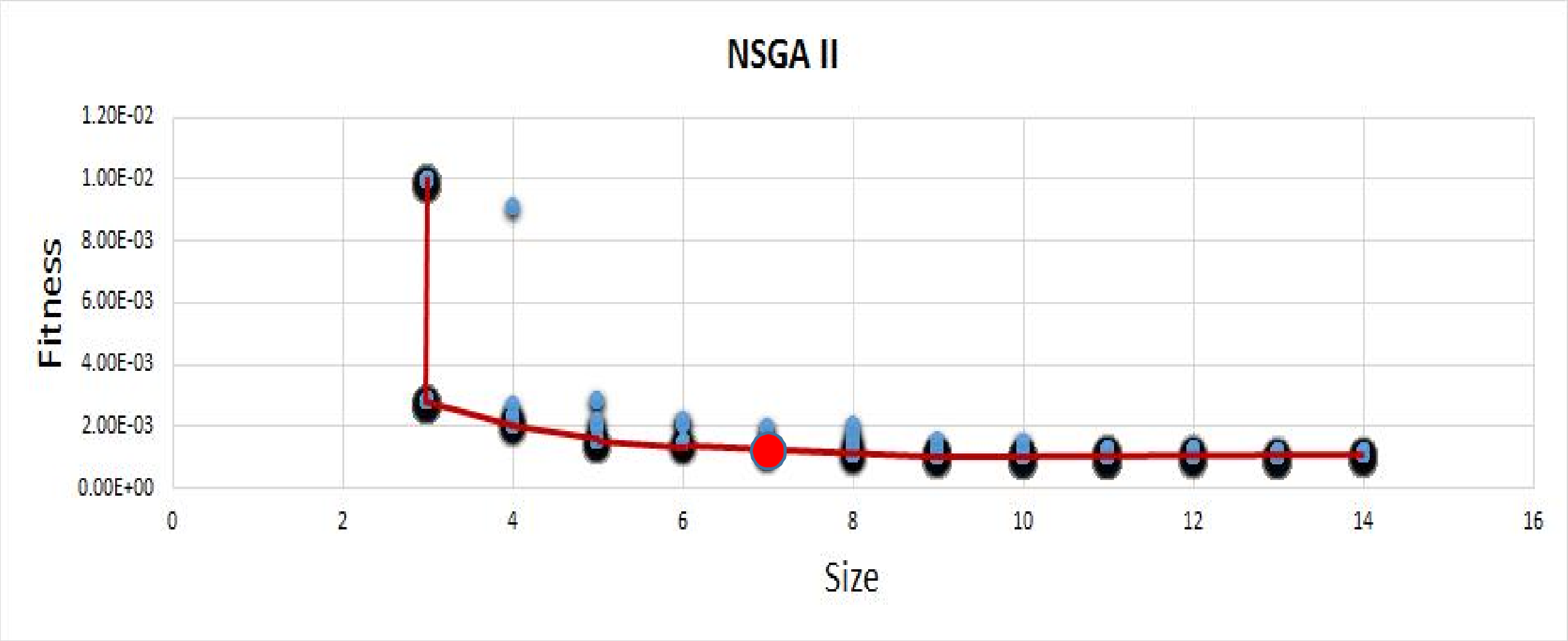

**Figure 1**: Optimal Pareto Front by NSGA II for Test Problem I

((cos((sin(C)*sqrt(cos(E)))))) ……………………[3]

Here the size of the solution is 7 and the fitness is 9.99 E-4. The solution is shown in dark circle on the pareto front in Figure 1.

Similarly executing using fitness function of SPEA 2 with elite size of 10, the obtained optimal front is shown in Figure 2. The additional parameters need to be specified for SPEA 2 are shown in the Table 7

**TABLE 7** : PARAMETERS IN SPEA 2

| Parameter in SPEA 2 | Changes in parameter file |
|---|---|
| Selection of parent parameter file | parent.0=../../../multiobjective/spea2/spea2.params |
| Usage of fitness function of SPEA 2 | pop.subpop.0.species.fitness = ec.multiobjective.spea2.SPEA2MultiObjectiveFitness |
| Selecting archive size | pop.subpop.0.archive-size =50 |
| Setting SPEA 2 Tournament Selection | pop.subpop.0.species.pipe.source.0.source.0 = ec.multiobjective.spea2.SPEA2TournamentSelection |
| Selection of Tournament Size | select.tournament.size= 2 |
| Setting elite size | breed.elite.0 = 10 |

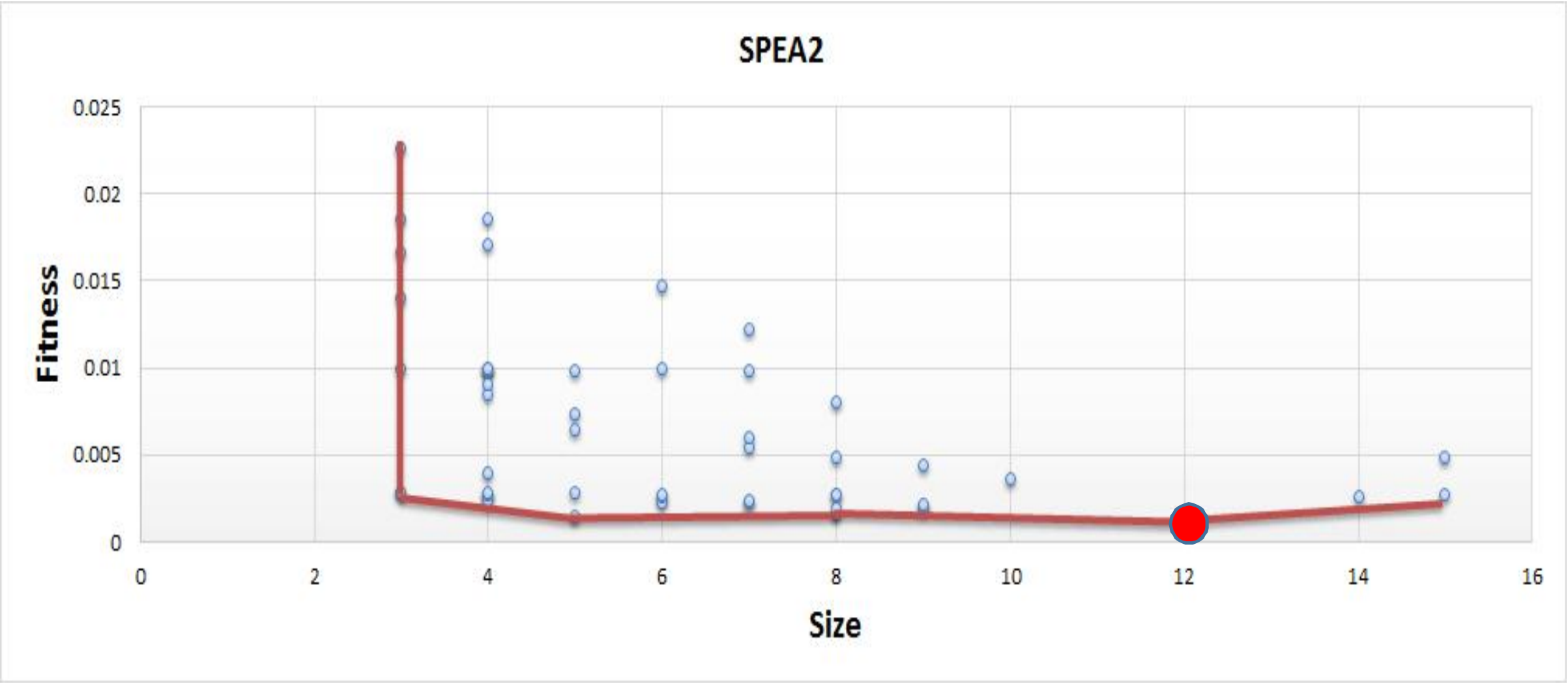

**Figure 2:** Optimal Pareto Front by SPEA II for Test Problem I

Thus the model generated by SPEA 2 is given below

((cos(sin(A))+(E*(B/7.0)))+sin((0.0*A)))  ………………………………..[4]

Here the size of the solution is 12 and the fitness is 0.0012. The bloat in the equation [sin(0.0*A)=0]is found which unnecessarily increases the size of equation by 4 which can be neglected. The solution is shown in dark circle on the pareto front in Figure 2.

Thus comparing equations [2] and [4] we can conclude that the size of the solution is reduced from 14 to 12 without affecting the fitness much.

*2) Test Problem II :*

Similarly implementing the experiments for the equation given below

y= sin(a) * (cos(b)/Sqrt(10^(c)) + tan(d-a)  …………..[5]

We have set number of genes equals to 5, gene head size equals to 8 and sub population size equals to 100 as shown in Table 8

**TABLE 8**:GENERAL PARAMETERS FOR TEST CASE II

| General Parameters | Changes in parameter file |
|---|---|
| Number of genes | gep.species.numgenes = 5 |
| Gene Head Size | gep.species.gene-headsize = 8 |
| Size of sub population | pop.subpop.0.size =100 |

Similarly for the Test Problem II the parameters for executing in GEP are same as in Table (2,3,4). The equation obtained by GEP is thus again compared to the equations obtained by executing the test problem under multi objective environment.

The equation thus obtained by GEP is given below

((((sin((0.0-D))+(B/exp((tan(sin(D))*exp(tan(B))))))+D)+(A-D))+(sin(tan(A))-B))  …………….[6]

The solution has size of 23 (removing the linking function) with the fitness of 0.001.

The parameters (Table 6) for NSGA II obtain results which are shown in Figure 3 and parameters (Table 7) for SPEA 2 whose results are shown in Figure 4.

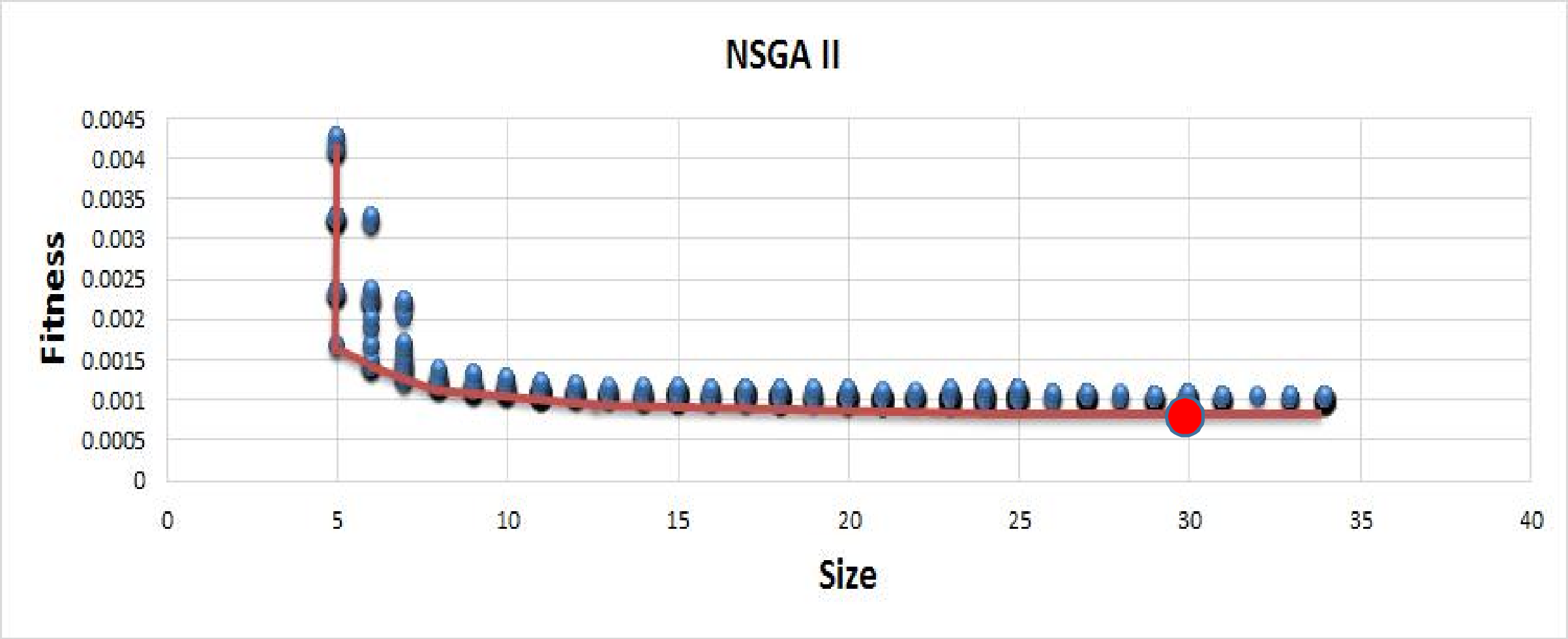

**Figure 3:** Optimal pareto front by NSGA II for *Test Problem II*

((((sin(sin(sin(sin(sin(tan(sqrt(3.0)))))))+sin(sin(sin((C/cos(A))))))+(C-exp(sin(D))))+cos(8.0))+cos(sin(sin(sin(tan(sin(tan(C)))))))) …[7]

Here the size of the solution is 30 and the fitness is 0.0010. The solution is shown in dark circle on the pareto front in Figure 3.

Similarly executing using fitness function of SPEA 2 with elite size of 10, the obtained optimal front is shown in Figure 4. Other parameters remain same as given in Table 7

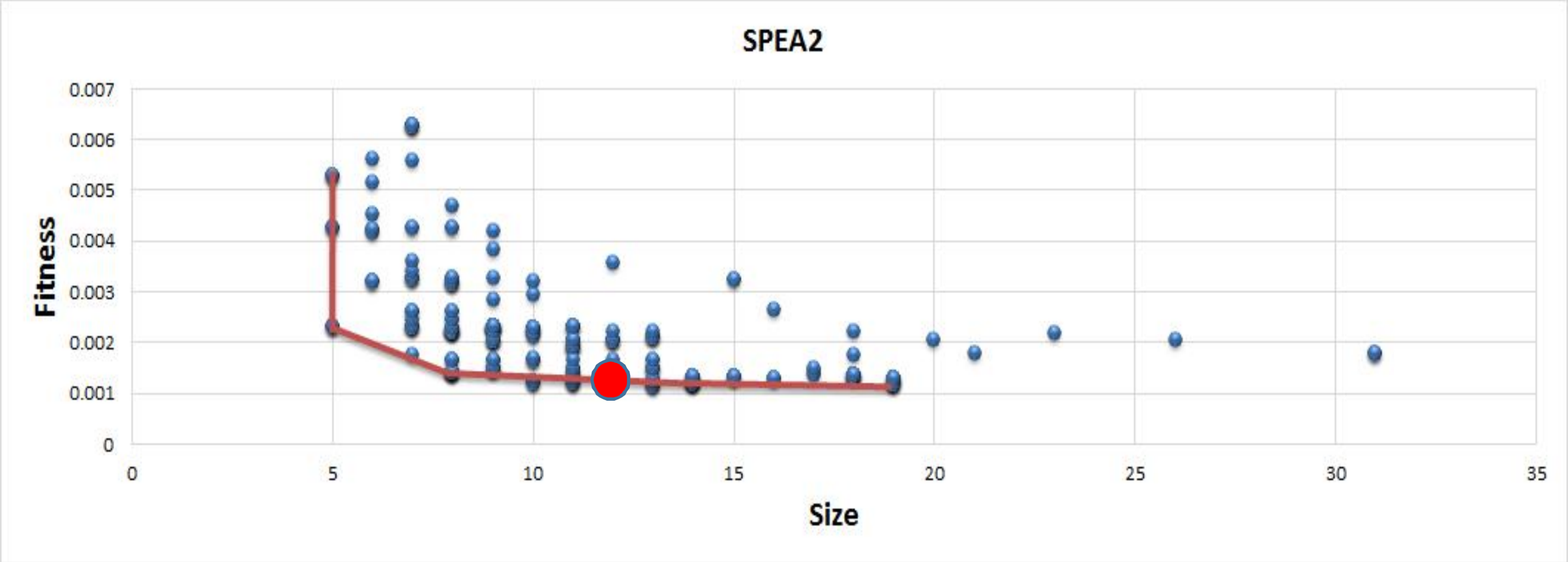

**Figure 4:** Optimal pareto front by SPEA II for Test Problem II

The equation obtained by SPEA 2 is shown below

((((cos(exp(B))+1.0)+B)+(cos(sin(6.0))-sqrt(7.0)))+sin(A)) ……………[8]

Here the size of the equation is 12 and the fitness is 0.0012 the solution is shown in dark circle on the pareto front in Figure 4.

Comparing equations [a2] and [b2] we can conclude that the size of the equation is reduced from 23 to 12 without affecting the fitness much.

*3) Dew Point*

We have considered series of dew point in which the independent variables are temperature (in °C) and relative humidity (in %) and dependant variables is dew point (in °C) [1][2][3]. The series contains monthly data ranging from (1st Janunary,1989 to 31st December,1993). The data was gathered from National Climatic Data Center(NCDC)(www.ncdc.noaa.gov) for New York City(Central Park).

The used function set is shown in Table 9

**TABLE 9**:FUNCTIONS USED

| Function Name | Weight | Symbol |
|---|---|---|
| Addition | 3 | + |
| Multiplication | 3 | * |
| Natural Logarithmic | 3 | Ln |
| Division | 3 | / |
| Exponential | 3 | Exp |

Other general parameters are shown in Table 10

**TABLE 10:** GENERAL PARAMETERS OF RUN

| Parameters | Value |
|---|---|
| Generations | 500 |
| Sub population size | 100 |
| Number of Chromosomes | 1 |
| Number of gene per Chromosome | 5 |
| Size of gene head | 4 |
| Fitness Function | RRSE(Root Relative Squared Error) |

We have used the constants in range of (-10) to (+10). Other general parameters for GEP are taken same as used in previous problems.

The final pareto front obtained using the fitness function of NSGA II and the parameters given in Table 6 is shown in Figure 5

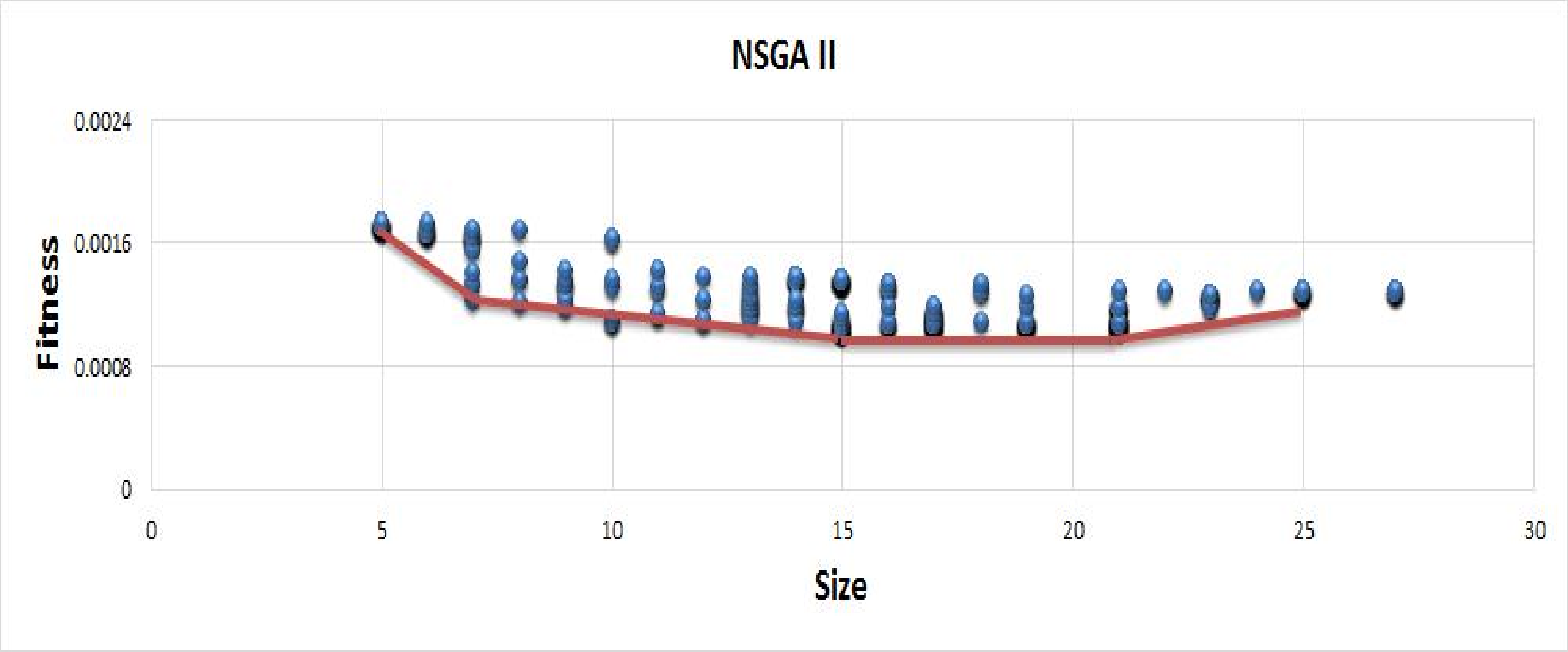

**Figure 5:**Optimal pareto front by NSGA II for *Dew Point*

(((((-4.688)+((-7.047)+((d0/d1)*(-7.047))))+((d1+d1)/8.773))+d0)+(-7.5)) ………………[9]

Using the same parameters for SPEA II as shown in Table (7) we get following results

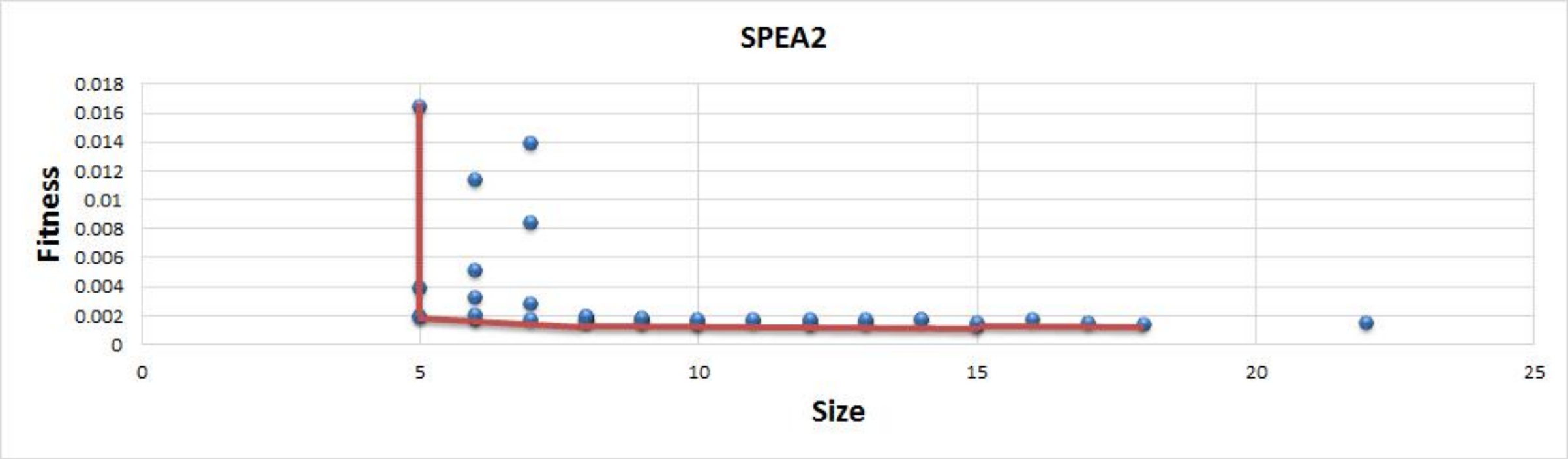

**Figure 6:** Optimal pareto front by SPEA II for *Dew Point*

## IV. COMPARISON OF RESULTS OBTAINED BY DEW POINT

Here d0 represents Temperature (ºC), d1 represents Relative Humidity (%) and D.V represents dew point (ºC)

Executing the Dew point application data using GEP evolves solution having fitness value of 0.001289. Here the linking function between two genes is ADD(+).

**Evolved GEP-MODEL in KARVA Notation**

Gene 0
d0.d0.C0.d0.d1.C1.C0.d0.d1
C0: 8.710252634681396
C1: -6.863296032971224
Gene 1
ln.d0./.exp.C1.C0.d1.C1.d1
C0: -7.621774506925439
C1: -2.0778512349549345
Gene 2
*.ln.C1.d0.C0.C1.d1.d1.C1
C0: -1.0538681881248397
C1: -9.381786492548889
Gene 3
/.d1.ln.d0.d0.C0.d0.d1.d0
C0: -0.2389522756341833
C1: -7.603029447944705
Gene 4
ln.exp./.d0.d1.d0.C0.d0.d0
C0: 0.5160341710891316
C1: 6.251276714801701

The simplified form of the solution is shown below:

**(d1+d0*ln(d0)+ln(d0)^2-9.381786492548889*ln(d0)^2+ln(d0)*ln(exp(d0/d1)))/ln(d0)** **……..**[10]

The error produced by the model over the original data shows that the mathematical model generated by Gene Expression Programming is not accurate. Moreover, the size of the obtained solution is 27 after removal of the linking function. We aim to produce accurate and smaller solution using multi objective based GEP.

Comparing the above results with the results obtained by multi objective algorithms like NSGA II and SPEA2 using the parameters specified in Table (6 & 7), the following results are obtained:-

***Worst Equation of the pareto front obtained using SPEA 2*** :-

The results obtained by SPEA 2 have its raw fitness value along with strength value S (the number of solutions it dominates or is dominated by) and the k-NN distance (D) from the nearest valued individual. The values in the below given fitness represent the two values of the objectives i.e fitness and the size of the equation. Suffix "min" at the end represents that both values are to be minimized to achieve optimal solution.

Fitness: [0.01640847 5.0 min]

S=13.0 D=0.19999863699062265 Final fitness: 0.19999863699062265
Linking function: Add

Thus evolved model is represented by each gene is shown below

**Evolved GEP-MODEL in KARVA Notation**
*Gene 0*
d0.d0.d0.exp.C1.d1.d1.d0.C0
C0: 3.490929536871512
C1: 4.358005672705509
*Gene 1*
d1.C1.C0.d1.C0.d0.C0.d0.d1
C0: -4.753958931516966
C1: -9.142275005578282
*Gene 2*
C1.d1.C0.exp.d0.d1.C1.d1.C1
C0: -5.323426761519379
C1: 0.13118796605988692
*Gene 3*
C1.d1./.d0.d0.C0.d0.C0.d1
C0: 9.103668908030528
C1: -0.304703823858933
*Gene 4*
d0.d1./.exp.d0.C1.C0.C1.d0
C0: 8.471428320162019
C1: 1.7920511892278448

The simplified form of model is

**0.13118796605988692-0.304703823858933+2*d0+d1 ……...……..[11]**

***Best equation of the pareto front obtained using SPEA 2:-***

Fitness: [0.0012465897 15.0 min]
S=28.0 D=0.5 Final fitness: 0.5
Linking function: Add

**Evolved GEP-MODEL in KARVA Notation**

*Gene 0*
/.d0.ln./.C0.d1.d0.d0.C0
C0: 5.303240984153685
C1: -7.737961811069729
*Gene 1*
/.d1.C0.d0.d1.C1.C0.C0.C0
C0: 6.431775793362785
C1: -9.216598283403313
*Gene 2*
ln.d0.d1.C1.d1.d1.d1.C0.d1

C0: 2.9305136571301773
C1: 8.465366760645846
*Gene 3*
C0.d1.*.d0.d0.d0.d0.d1.d1
C0: -4.883511342816025
C1: 2.137540064728567
*Gene 4*
+.C1.d0.+.d0.d0.d0.d0.d1
C0: 1.5462946814544747
C1: -5.293498296641729

The simplified form of the equation gives

**(((((d0/ln((5.303240984153685/d1)))+(d1/6.431775793362785))+ln(d0))+(-4.883511342816025))+(d0+(-5.293498296641729))) ………………[12]**

Comparing the best solutions obtained using GEP and SPEA 2 based GEP, we noticed that GEP evolved solution has size 27 with fitness value of 0.001289, whereas SPEA 2 based GEP generated solution with the size of 15 and fitness value of 0.001247.

Figure presents the observed /actual dew point series generated using GEP generated solution , series generated using SPEA 2 based GEP generated solution and difference between these actual series and modelled series.

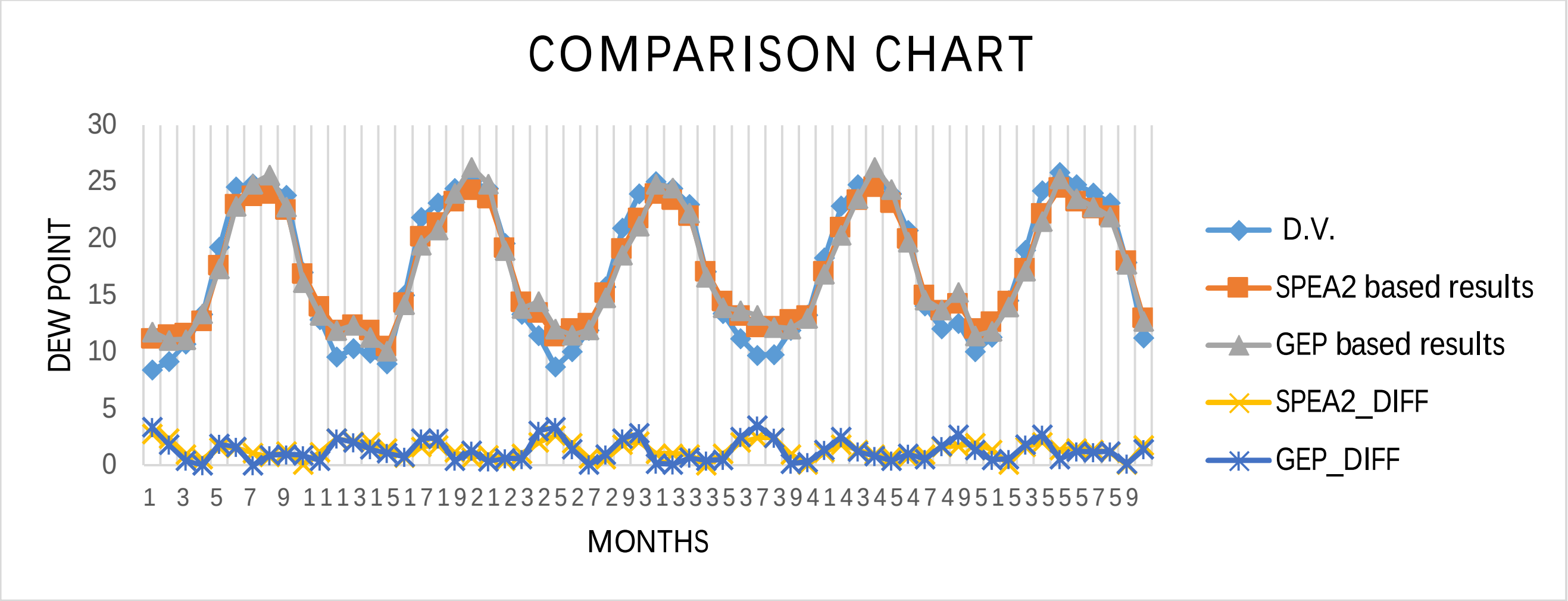

**Figure 7: Comparison of results obtained by different models to the original values**

Prediction is done with the basic idea of using 75 % of the data in training and then predicting rest 25 % of data with the models. As specified earlier the training data used for modelling the dew point is monthly data ranging from year 1989 to 1993. The testing data ranging from year 1993 to 1995 is used to test the models obtained by GEP and GEP along with SPEA 2. The below shown graph represents the prediction of dew point.

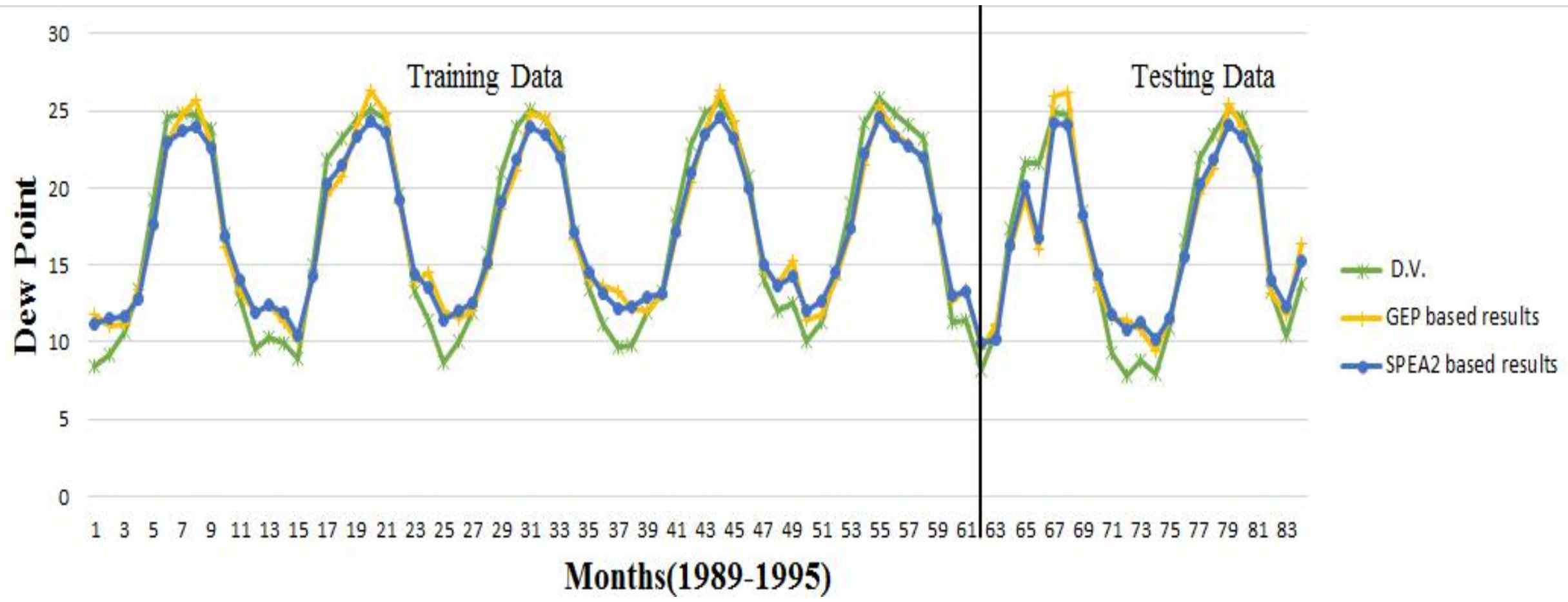

**Figure 8:Prediction of Dew Point**

Thus experiments on GEP in multi objective environment gives promising results considering the trade-off between multiple objectives.

## V. Conclusions

After survey of different modeling techniques and analysis of their pros and cons we came to conclusion that Gene Expression Programming is better for data base modelling due to it does not assume model structure a priori. The solutions generated by GEP are accurate enough but larger in size. This restricts the applicability of the generated solutions. Multi-objectives optimization algorithms like NSGA and SPEA are used for generating solutions that are smaller in size with higher accuracy. We have used ECJ toolkit for GEP based and multi-objective based GEP modelling. Two test problems are modeled using GEP and GEP based multi objective optimization. The results on test problems, suggests use of SPEA 2 because it returns the least size of optimal pareto front. The experiments performed for Dew Point data series under multi-objective optimization using SPEA2 generates smaller and accurate solutions as compared to the solutions generated by GEP.

## References

[1] Lawrence, M. G. (2005). The Relationship between Relative Humidity and the Dewpoint Temperature in Moist Air: A Simple Conversion and Applications. Bulletin of the American Meteorological Society, 86(2), 225–233. doi:10.1175/BAMS-86-2-225

[2] Andrei, B., & Elena, B. (2006). Meteorological Data Analysis and Prediction by Means of Genetic Programming, 35–42.

[3] Bărbulescu, A., & Băutu, E. (2009). Mathematical models of climate evolution in Dobrudja. Theoretical and Applied Climatology, 100(1-2), 29–44. doi:10.1007/s00704-009-0160-7

[4] Neill, O., Agapitos, A., Neill, M. O., & Brabazon, A. (2012). Genetic Programming for the Induction of Seasonal Forecasts : A Study on Weather-derivatives Author ( s ) Financial Decision Making using Computational Intelligence , Series in Optimisation and its Applications Publisher Genetic Programming for the Induction of Seasonal Forecasts A Study on Weather-derivatives.

[5] Ferreira, C. (2001). Gene Expression Programming in Problem Solving, (1992).

[6] Iberian, X. X. V, American, L., & Methods, C. (2004). A GENE EXPRESSION PROGRAMMING SYSTEM, 10–12.

[7] Bakhshaii, A., & Stull, R. (2009). Deterministic Ensemble Forecasts Using Gene-Expression Programming*. Weather and Forecasting, 24(5), 1431–1451. doi:10.1175/2009WAF2222192.1

[8] Ferreira, C. (2001). Gene Expression Programming : A New Adaptive Algorithm for Solving Problems, 1–22.

[9] Taylor, P., Parasuraman, K., Elshorbagy, A., & Carey, S. K. (n.d.). Modelling the dynamics of the evapotranspiration process using genetic programming Modelling the dynamics of the evapotranspiration process using genetic programming, (September 2012), 37–41.

[10] Konak, A., Coit, D. W., & Smith, A. E. (2006). Multi-objective optimization using genetic algorithms: A tutorial. Reliability Engineering & System Safety, 91(9), 992–1007. doi:10.1016/j.ress.2005.11.018

[11] Kalyanmoy Deb .Single and Multi Objective optimization using Evolutionary Algorithm

[12] Zitzler, E., Deb, K., & Thiele, L. (1999). Comparison of Multiobjective Evolutionary Algorithms : Empirical Results 1 Motivation.

[13] Zitzler, E., Laumanns, M., & Thiele, L. (2001). SPEA2 : Improving the Strength Pareto Evolutionary Algorithm, 1–21.

[14] Zitzler, E., & Thiele, L. (1999). Multiobjective Evolutionary Algorithms : A Comparative Case Study and the Strength Pareto Approach, 3(4), 257–271.

[15] Deb, K., Member, A., Pratap, A., Agarwal, S., & Meyarivan, T. (2002). A Fast and Elitist Multiobjective Genetic Algorithm :, 6(2), 182–197.

[16] Corne, David,Knowles Joshua,Oates Martin. The Pareto Envelope Based selection Algorithm for Multi Objective Optimization.

[17] Ngatchou, P., & Zarei, A. (2005). Pareto Multi Objective Optimization, 84–91.

[18] Sbalzarini, B. I. F., Sibylle, M., & Koumoutsakos, P. (2000). Multiobjective optimization using evolutionary algorithms.
[19] Jie, Z., Changjie, T., Chuan, L., Anlong, C., & Chang, Y. (n.d.). Time Series Prediction based on Gene Expression Programming, (1).
[20] O'Neill, M., Vanneschi, L., Gustafson, S., & Banzhaf, W. (2010). Open issues in genetic programming. Genetic Programming and Evolvable Machines, 11(3-4), 339–363. doi:10.1007/s10710-010-9113-2
[21] Luke, S. (n.d.). Two Fast Tree-Creation Algorithms for Genetic Programming, 1–9.
[22] Luke, S. (1996). A Survey and Comparison of Tree Generation Algorithms.
[23] Eggermont, J. (1859). Genetic Programming.
[24] Savic, D. A., Walters, G. A., & Davidson, J. W. (1999). A Genetic Programming Approach to Rainfall-Runoff Modelling, 219–231.